\def\paperID{X}
\def\confName{CVPR}
\def\confYear{2025}

\def\paperTitle{GFreeDet: Exploiting Gaussian Splatting and Foundation Models for\\Model-free Unseen Object Detection in the BOP Challenge 2024}

\def\authorBlock{
    Xingyu Liu$^{*}$,~~~
    Gu Wang$^{*,\dagger}$,~~~
    Chengxi Li$^{*}$,~~~ 
    Yingyue Li$^{*}$,~~~ 
    Chenyangguang Zhang,~~~\\
    Ziqin Huang,~~~
    and Xiangyang Ji$^{\dagger}$ \\
    Tsinghua University \\
    {\tt\small \{liuxy21, lichengx21, yingyue-21, zcyg22, huang-zq24\}@mails.tsinghua.edu.cn} \\
    {\tt\small \{wanggu1, xyji\}@tsinghua.edu.cn}
}

\newif\ifreview 
\newif\ifarxiv \newcommand{\arxiv}{\arxivtrue}
\newif\ifcamera 
\newif\ifrebuttal 

\arxiv

\pdfoutput=1
\documentclass[10pt,twocolumn,letterpaper]{article}
\ifreview \usepackage[review]{cvpr} \fi
\ifarxiv \usepackage[pagenumbers]{cvpr} \fi
\ifrebuttal \usepackage[rebuttal]{cvpr} \fi
\ifcamera \usepackage{cvpr} \fi

\usepackage{graphicx}	
\usepackage{amsmath}	
\usepackage{amssymb}	
\usepackage{booktabs}
\usepackage{times}
\usepackage{microtype}
\usepackage{epsfig}
\usepackage{caption}
\usepackage{float}
\usepackage{placeins}
\usepackage{color, colortbl}
\usepackage{stfloats}
\usepackage{enumitem}
\usepackage{tabularx}
\usepackage{xstring}
\usepackage{multirow}
\usepackage{xspace}
\usepackage{url}
\usepackage{subcaption}
\usepackage{xcolor}
\usepackage[hang,flushmargin]{footmisc}

\ifcamera \usepackage[accsupp]{axessibility} \fi

% ...

\newcolumntype{L}[1]{>{\raggedright\let\newline\\\arraybackslash\hspace{0pt}}m{#1}}
\newcolumntype{C}[1]{>{\centering\let\newline\\\arraybackslash\hspace{0pt}}m{#1}}
\newcolumntype{R}[1]{>{\raggedleft\let\newline\\\arraybackslash\hspace{0pt}}m{#1}}

\newlength\savewidth\newcommand\shline{\noalign{\global\savewidth\arrayrulewidth
  \global\arrayrulewidth 1pt}\hline\noalign{\global\arrayrulewidth\savewidth}}
\newcommand{\tablestyle}[2]{\setlength{\tabcolsep}{#1}\renewcommand{\arraystretch}{#2}\centering\footnotesize}

\ifarxiv  \fi

\newcommand{\R}[1]{{%
    \textbf{%
        \ifstrequal{#1}{1}{\textcolor{red}{R#1}}{%
        \ifstrequal{#1}{2}{\textcolor{blue}{R#1}}{%
        \ifstrequal{#1}{3}{\textcolor{magenta}{R#1}}{%
        \ifstrequal{#1}{4}{\textcolor{teal}{R#1}}{%
                           \textcolor{cyan}{R#1}%
        }}}}%
    }%
}}

\newcommand{\cellOne}{{\cellcolor{green!30}}}
\newcommand{\cellTwo}{{\cellcolor{yellow!30}}}
\newcommand{\cellThree}{{\cellcolor{orange!30}}}

\newcommand{\BgOne}[1]{{\colorbox{green!30}{#1}}}
\newcommand{\BgTwo}[1]{{\colorbox{yellow!30}{#1}}}
\newcommand{\BgThree}[1]{{\colorbox{orange!30}{#1}}}  

\usepackage{xr-hyper}

\makeatletter
\newcommand*{\addFileDependency}[1]{
  \typeout{(#1)}
  \@addtofilelist{#1}
  \IfFileExists{#1}{}{\typeout{No file #1.}}
}

\makeatother
\newcommand*{\myexternaldocument}[1]{
    \externaldocument{#1}
    \addFileDependency{#1.tex}
    \addFileDependency{#1.aux}
}

\definecolor{cvprblue}{rgb}{0.21,0.49,0.74}
\usepackage[pagebackref,breaklinks,colorlinks,allcolors=cvprblue]{hyperref}
\usepackage[capitalize]{cleveref}
\crefname{section}{Sec.}{Secs.}
\crefname{table}{Table}{Tables}
\crefname{figure}{Fig.}{Figs.}

\ifarxiv \crefname{appendix}{App.}{Apps.}
\else \crefname{appendix}{Suppl.}{Suppls.} \fi

\unless\ifarxiv \myexternaldocument{_supplementary} \fi

\begin{document}
\title{\paperTitle}
\author{\authorBlock}
\maketitle

\begin{abstract}
We present \underline{GFreeDet}, an unseen object detection approach that leverages \underline{G}aussian splatting and vision \underline{F}oundation models under model-f\underline{ree} setting. Unlike existing methods that rely on predefined CAD templates, GFreeDet reconstructs objects directly from reference videos using Gaussian splatting, enabling robust detection of novel objects without prior 3D models. Evaluated on the BOP-H3 benchmark, GFreeDet achieves comparable performance to CAD-based methods, demonstrating the viability of model-free detection for mixed reality (MR) applications. Notably, GFreeDet won the best overall method and the best fast method awards in the model-free 2D detection track at BOP Challenge 2024.
\let\thefootnote\relax\footnotetext{$^{*}$ Equal contribution.}
\let\thefootnote\relax\footnotetext{$^{\dagger}$ Corresponding authors.}
\end{abstract}

\section{Method and Technical Details}
\label{sec:method}

\subsection{Problem Statement} 
\label{sec:problem_def}
Model-free unseen object detection \footnote{\href{https://bop.felk.cvut.cz/tasks/\#ModelFree-2DDet-Unseen}{bop.felk.cvut.cz/tasks/\#ModelFree-2DDet-Unseen}} is a newly introduced task in BOP Challenge 2024 \cite{nguyen2025bopchallenge2024modelbased}\footnote{\href{https://bop.felk.cvut.cz/challenges/bop-challenge-2024}{bop.felk.cvut.cz/challenges/bop-challenge-2024}}.
To mitigate the limitation of model-based tasks for open-world scenarios where CAD models are often not available, the model-free setting opts for rapidly learning unseen objects just from video(s).   
To facilitate this, a short onboarding stage (\ie, a maximum of 5 minutes on 1 GPU) is given to a method for each new object, which involves learning from reference video(s) showing all possible views at the object.

In this work, we focus on the static onboarding scenario, where two reference videos are provided: one captures the object in an upright pose, and the other in an upside-down pose, both standing still on a desk, displaying all possible views.
Ground-truth object poses are available for all frames, as methods like COLMAP \cite{schoenberger2016sfm} can relatively easily offer very accurate poses in such scenarios.
During inference, the method is given an RGB or grayscale image unseen that shows an arbitrary number of instances of an arbitrary number of test objects, with all objects being from one specified dataset of the BOP-H3 datasets (HOT3D \cite{banerjee2024hot3d}, HOPEv2 \cite{tyree2022hope}, HANDAL \cite{handaliros23_handal}). 
No prior information about the visible object instances is provided.
Following the 2D detection tasks of previous BOP challenges \cite{Hodan_2024_CVPR}, the goal is to produce a list of object detections with confidences for each test image, with each detection defined by an amodal 2D bounding box.

\begin{figure*}[t]
    \centering
    \includegraphics[width=\linewidth]{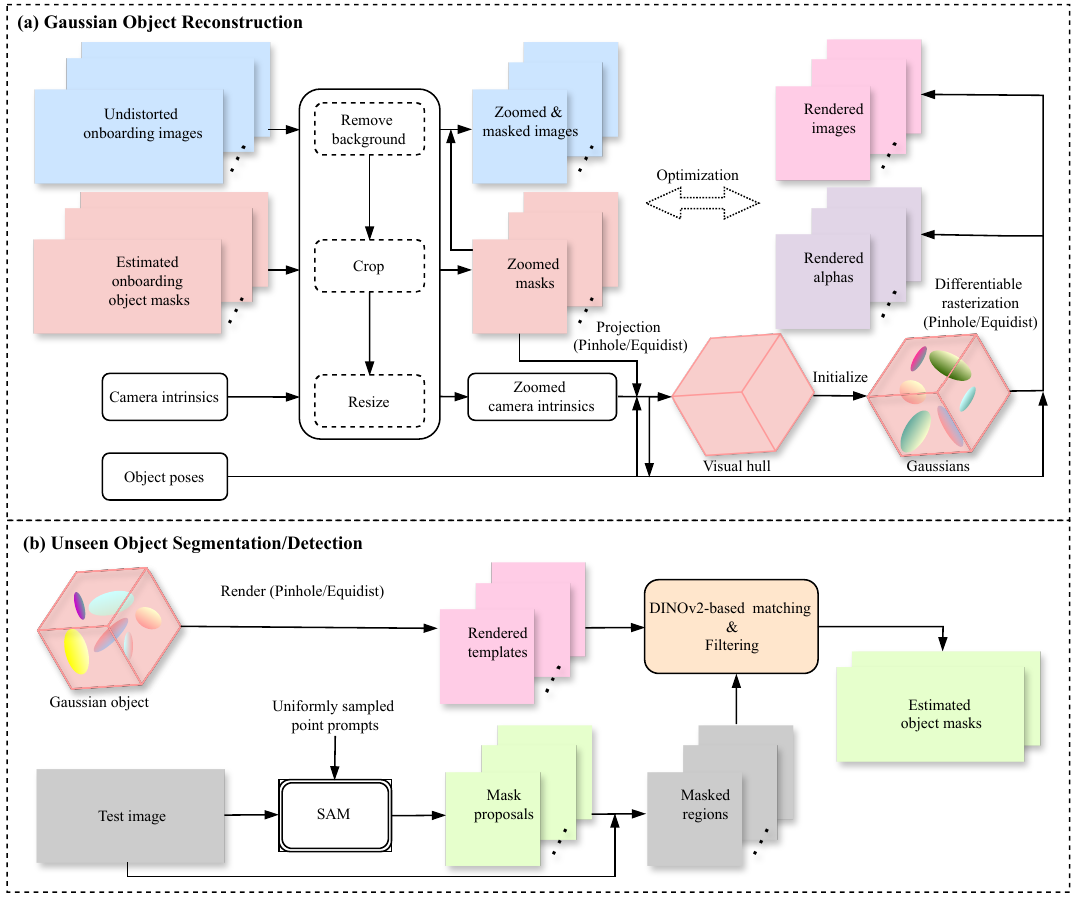}
    \caption{Overview of the pipeline of GFreeDet for model-free unseen object detection. For an unseen object, we first reconstruct the Gaussian object given calibrated onboarding images, object poses, and the estimated object masks. The Gaussian object is then used to render templates. During inference, we leverage DINOv2 to match the masked regions obtained by SAM against the rendered templates. The final object masks are obtained after filtering the matched results.}
    \label{fig:pipeline}
\end{figure*}

\subsection{Overview of GFreeDet}
\label{sec:method_overview}
Our method for model-free unseen object detection, which we dub GFreeDet, is built on top of Gaussian splatting \cite{tog23_3dgs} and vision foundation models such as the segment anything model (SAM) \cite{kirillov2023sam} and DINOv2 \cite{oquab2023dinov2}. 
An overview of the GFreeDet pipeline is depicted in Fig.~\ref{fig:pipeline}. 

During the onboarding stage, we reconstruct a Gaussian object for the unseen object based on the posed static onboarding frames, which is then used to render templates.
During inference, considering the zero-shot generalizability of SAM, we cast the detection task as an instance segmentation task as the bounding boxes can be easily obtained from the instance-level masks.
Thereby, we leverage SAM and DINOv2 to obtain all possible instance masks for all possible objects defined in a given dataset through matching the mask proposal regions with the rendered templates.

Note that except for the training of the Gaussian object during the onboarding stage, our approach does not need any additional training.

\subsection{Gaussian Object Reconstruction}
\label{sec:go}

Given static onboarding images, along with camera intrinsics, ground-truth object poses, and the estimated object masks, we leverage Gaussian splatting \cite{tog23_3dgs} to reconstruct unseen objects.

We first utilize the estimated onboarding object masks to extract the target object out of the background.
Subsequently, we crop and resize the images to a fixed size using these masks, resulting in zoomed and masked images. 
This process not only speeds up the training of Gaussian splatting but also improves the quality of the reconstructed Gaussian object.
Afterwards, we build the 3D visual hull \cite{yang2024gaussianobject} from the zoomed masks to effectively initialize the positions of Gaussian primitives. 
Essentially, the visual hull is derived by retaining only the valid 3D points whose corresponding projected 2D positions fall within the foreground object regions.

For the training of Gaussian splatting, we employ a weighted combination of the L1 loss and the SSIM loss~\cite{Wang04_ssim} between the rendered images and ground-truth onboarding images, supplemented by a silhouette loss between the rendered alphas and the zoomed object masks.

Noteworthy, we unify our pipeline for both images captured with both pinhole cameras and fisheye cameras.
For each type of camera, the images are first undistorted with the given distortion parameters without changing the camera projection method. 
Then we use the corresponding projection method, \ie, perspective projection for the pinhole camera and equidist projection for the fisheye camera, to build the visual hull and rasterize Gaussians.
Specifically, we employ the Gaussian rasterization method implemented in gsplat~\cite{ye2025gsplat} which supports both pinhole projection and equidist projection \cite{liao2024fisheyegs}. 

\begin{figure}[!t]
    \centering
    \includegraphics[width=\linewidth]{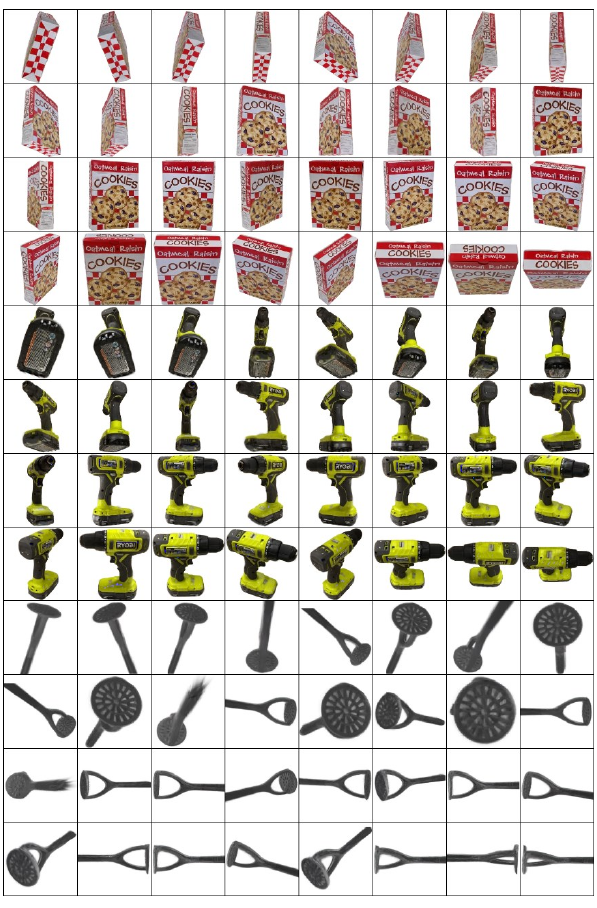}
    \caption{Visualization of reconstructed templates rendered by Gaussian Splatting. The objects are selected from HOPEv2, HANDAL and HOT3D from top to bottom.}
    \vspace{-1mm}
    \label{fig:temp}
\end{figure}

\subsection{Unseen Object Detection}
\label{sec:seg}

\paragraph{Template Descriptors}  Given the Gaussian model, we render $N_\mathcal{T}$ RGB-D synthetic templates covering different orientations of the object using Gaussian splatting as shown in Fig.~\ref{fig:temp}. 
Here we set $N_\mathcal{T}$ to 162 following SAM-6D~\cite{lin2023sam6d}.
For each template $t$, we generate an image-level global descriptor $\mathcal{G}_{t}$ and $N_l$ patch-level descriptors $\{\mathcal{L}_{t}^j\}_{j=1}^{N_l}$ using DINOv2~\cite{oquab2023dinov2}.

\paragraph{Mask Proposal Descriptors}
For each testing image, we use SAM~\cite{kirillov2023sam} or Fast-SAM~\cite{zhao2023fastsam} to generate $N_{\mathcal{P}}$ mask proposals, where uniformly sampled pixel positions are used as prompt.
Subsequently, we mask and crop the test image with all mask proposals and resize them to a consistent size $\{I_p \in \mathbb{R}^{224 \times 224 \times 3}\}_{p=1}^{N_{\mathcal{P}}}$.
Similarly, for each unlabeled image proposal $I_{p}$, we generate an image-level global descriptor $\mathcal{G}_{p}$ and $N_l$ patch-level descriptors $\{\mathcal{L}_{p}^j\}_{j=1}^{N_l}$ using DINOv2~\cite{oquab2023dinov2}.

\paragraph{Template Matching}
In the matching stage, we aim to assign each mask proposal an object category and a matching score.
Specifically, given the mask proposal $I_{p}$, we first compute a global score by averaging the top $K$ values of cosine similarity between template global descriptors ($K = 5$), \ie,
\begin{equation}
    s_{\mathcal{G}, t} = \text{Avg}_\text{topK} \frac{\mathcal{G}_{p}^\top \mathcal{G}_{t}}{\|\mathcal{G}_{p}\|_2 \cdot \|\mathcal{G}_{t}\|_2 }.
\end{equation}
The category of this mask proposal is defined by the template set with the highest global score, denoted as $\overline{t}$.

After identifying the category, we further compute a local matching score by
\begin{equation}
    s_\mathcal{L} =  \frac{1}{N_l} \sum_{k=1}^{N_l} 
\max_{i=1,...,N_l}
\frac{ (\mathcal{L}_p^k)^\top \mathcal{L}_{\overline{t}}^i}{\|\mathcal{L}_p^k\|_2 \cdot \|\mathcal{L}_{\overline{t}}^i\|_2}.
\end{equation}
The overall matching score is computed as
\begin{equation}
    s = \frac{s_{\mathcal{G}, \overline{t}}+s_\mathcal{L}}{2}.
\end{equation}
Harnessing both global and local descriptors, the method effectively assigns each mask proposal a category and matching score.

Finally, we obtain the estimated masks by filtering out low-score predictions and eliminating duplicates using Non-Maximum Suppression.
The bounding boxes are calculated based on the tightest fit around the object masks.

Compared to other datasets, the HOT3D dataset poses specific challenges for the task of detecting unseen objects. 
One notable difference is that the onboarding data is exclusively in grayscale, whereas the test data encompasses both RGB and grayscale images. 
Furthermore, unlike other datasets that are captured using pinhole cameras, HOT3D utilizes fisheye cameras for all its images. 
To streamline the process, we convert the RGB test images to grayscale.
Additionally, the templates for HOT3D are rendered using a fisheye camera model, specifically employing an equidistant projection with zero distortion parameters.

\subsection{Implementation Details}
\label{sec:impl}

We implemented our approach with PyTorch \cite{paszke2019pytorch}. 
All the experiments were conducted on a computer equipped with an RTX 3090 GPU.
All the onboarding images were first cropped and resized to a resolution of 256$\times$256. 
For constructing visual hulls, we utilized the farthest point sampling (FPS) \cite{eldar1997fps} strategy based on the object rotations to select up to 8 viewpoints from each onboarding sequence.
For the training of Gaussian splatting, we used the whole onboarding sequence as input viewpoints, including object captured in both upright and upside-down poses.
The maximum spherical harmonic (SH) degree of Gaussians was set to 2.
We trained each Gaussian object for 10K iterations, during which the opacities were reset every 1K iterations and density control was disabled after 6K iterations.
\section{Experimental Results}
\label{sec:exp_results}

\begin{table*}[t]
\centering
\tablestyle{7pt}{1.3}
\begin{tabular}{l lc |ccc |c c}
Method & Seg. Model & Onboarding Type & AP$_\text{HOT3D}$ (\%) & AP$_\text{HOPEV2}$ (\%) & AP$_\text{HANDAL}$ (\%) & AP$_\text{H3}$ (\%) & Time (s) \\
\shline
MUSE & Unknown & CAD & \cellOne 42.6 & \cellOne 47.4 & \cellOne 27.0 & \cellOne 39.0 & \cellTwo 1.485 \\
CNOS \cite{nguyen2023cnos}& FastSAM \cite{zhao2023fastsam} & CAD & \cellTwo 35.0 & \cellThree 31.3 & \cellTwo 24.6 & \cellTwo 30.3 & \cellOne 0.332 \\
CNOS \cite{nguyen2023cnos}& SAM \cite{kirillov2023sam} & CAD & \cellThree 31.7 & \cellTwo 36.5 & \cellThree19.7 & \cellThree 29.3 & \cellThree 1.786 \\
\hline
GFreeDet (Ours) & FastSAM \cite{zhao2023fastsam} & Static & \cellOne 33.8 & \cellTwo 36.4 & \cellTwo 25.5 & \cellOne 31.9 & \cellOne 0.278 \\
GFreeDet (Ours) & SAM \cite{kirillov2023sam}& Static & \cellTwo 30.9 & \cellOne 38.4 & \cellOne 26.4 & \cellOne 31.9 & \cellTwo 2.136 \\
CNOS \cite{nguyen2023cnos}&	SAM \cite{kirillov2023sam}&	Static&	-&	\cellThree 34.5&	-	&-	&-\\
\end{tabular}
\caption{\label{tab:results}
Comparison of different unseen object detection methods on the BOP-H3 datasets. We report the AP (\%) on each dataset, the average AP (\%) on the BOP-H3 datasets, and the average time per image (s). All the results are available on \href{https://bop.felk.cvut.cz/leaderboards}{bop.felk.cvut.cz/leaderboards}. Best results for each onboarding type are highlighted as \BgOne{1st}, \BgTwo{2nd}, and \BgThree{3rd}.
}
\end{table*}

\subsection{Datasets and Evaluation Metrics}
\label{sec:data_metric}

\paragraph{Datasets} 
The model-free unseen object detection task of BOP Challenge 2024 is evaluated on the BOP-H3 datasets, which consist of three datasets: HOT3D \cite{banerjee2024hot3d}, HOPEv2 \cite{tyree2022hope}, and HANDAL \cite{handaliros23_handal}.

The HOT3D dataset \cite{banerjee2024hot3d} is designed for 3D egocentric hand and object tracking, offering multi-view perspectives with RGB and monochrome fisheye image streams. It captures 19 subjects engaging with 33 distinct rigid objects.
It not only includes basic actions such as pick-up/observe/put-down objects but also encompasses activities that mimic daily tasks found in kitchen, office, and living room settings. 
It is recorded by two Meta head-mounted devices: Project Aria, a light-weight research prototype for AR/AI glasses, and Quest 3, a production VR headset. 
For the BOP Challenge 2024, the static onboarding videos of the HOT3D dataset are captured using the Quest 3, which is equipped with monochrome fisheye cameras. 
The test set comprises 5140 frames across 1028 scenes, recorded with both Project Aria and Quest 3 devices. 
The detection results are assessed based on the left grayscale image from Quest 3 and the RGB image from Project Aria.

HOPEv2, the upgraded version of the HOPE dataset~\cite{tyree2022hope}, is newly introduced in the BOP Challenge 2024. 
This dataset features 28 toy grocery objects photographed within household and office settings. 
It expands upon the original HOPE dataset by maintaining the initial 40 test scenes and supplementing them with 7 additional scenes from seven cluttered environments. 
The original 40 scenes were captured across 8 diverse environments, each with 5 configurations, yielding a total of 40 distinct scenes characterized by their camera and object poses. 
In contrast, the newly integrated cluttered scenes are captured by a moving Vicon camera.
As a result, HOPEv2 offers 457 comprehensive test images for the BOP Challenge 2024.

The HANDAL dataset \cite{handaliros23_handal} comprises graspable or manipulable objects like hammers, ladles, cups, and power drills, captured from various views within cluttered, multi-object scenarios. 
For the BOP Challenge 2024, additional testing images were introduced. Unlike the original dataset which features 212 objects across 17 categories, the BOP Challenge 2024 focuses on a subset of 40 objects from 7 categories. 
As a resut, the BOP test set encompasses 6643 images from 61 distinct scenes.

\begin{figure*}[!th]
    \centering
    \includegraphics[width=\linewidth]{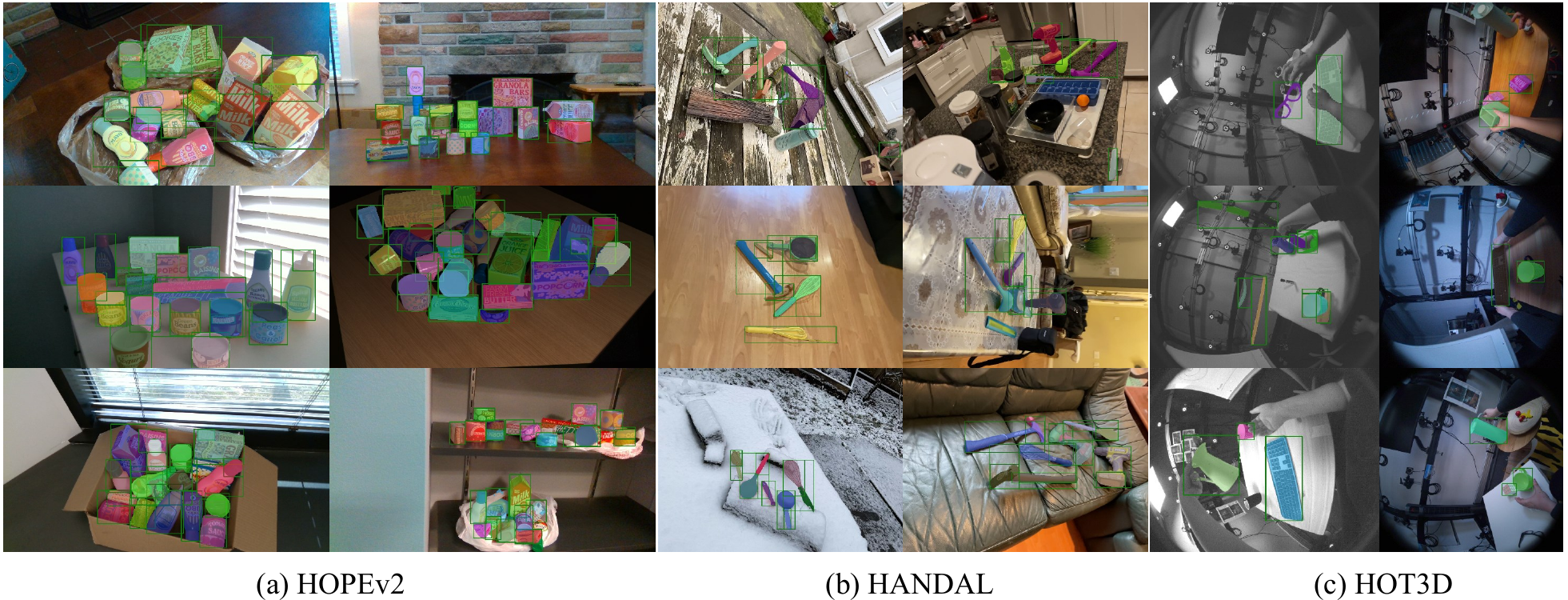}
    \caption{Qualitative results on BOP-H3 datasets. Objects are colored by predicted masks. Best viewed by zooming in.}
    \label{fig:res}
\end{figure*}

\paragraph{Evaluation Metrics} 
The BOP Challenge assesses object detection accuracy using a modified version of the Average Precision (AP) metric from the COCO Object Detection Challenge \cite{lin2014microsoft}.
Concretely, the AP score for each object is determined by calculating the average precision across a range of IoU (Intersection over Union) thresholds, \ie, [0.5, 0.55, ..., 0.95].
Distinct from the evaluation method in COCO, BOP's evaluation ignores predictions for annotated objects that are nearly invisible (less than 10\% visibility).
The dataset-level AP is obtained by averaging the per-object AP scores of all objects within the given dataset.
Specifically, for the unseen object detection tasks on BOP-H3 datasets, there are three dataset-level APs: AP$_\text{HOT3D}$, AP$_\text{HOPEv2}$, and AP$_\text{HANDAL}$. 
Furthermore, an overall metric AP$_\text{H3}$ is derived by averaging these three dataset-level APs, providing a comprehensive measure of performance across the three datasets.

Besides the AP scores, the average processing time per image is also employed as a metric because the detection speed is essential for real-world applications.

\subsection{Results}
\label{sec:results}

The results of different unseen object detection methods on the BOP-H3 datasets are presented in Tab.~\ref{tab:results} and Fig.~\ref{fig:res}.
All the results are obtained from the public leaderboard of the BOP Challenge 2024.
We compare with all the publicly available methods submitted to both the model-based and model-free unseen object detection tracks until the deadline of the BOP Challenge 2024.
For all methods, we report results in terms of the AP (\%) on each of the BOP-H3 datasets, the average AP (\%) on BOP-H3, and the average processing time per image (s).

As evidenced by the superior results in Tab.~\ref{tab:results}, the performance of our GFreeDet, utilizing either SAM or FastSAM, stands out in the model-free track for unseen object detection given static onboarding images.
Both GFreeDet-SAM and GFreeDet-FastSAM achieve an AP$_\text{H3}$ of 31.9\,\% on the BOP-H3 datasets, with GFreeDet-FastSAM demonstrating a significant speed advantage.
In comparison, the prior-art CNOS~\cite{nguyen2023cnos}, which only reported results on HOPEv2, falls short in terms of AP$_\text{HOPEv2}$ when measured against our GFreeDet.
Notably, our GFreeDet even surpasses CNOS with CAD-based onboarding in terms of AP$_\text{H3}$, highlighting the great potential of model-free approaches in detecting unseen objects.
Despite this, the submitted method MUSE, which employs CAD-based onboarding data, still leads in AP$_\text{H3}$ among all the compared methods, although it has yet to disclose its implementation details.
Nonetheless, our GFreeDet-FastSAM not only boasts the quickest inference speed among all competitors but also delivers an elegant detection AP, excelling in both aspects regardless of the onboarding type.
Consequently, our GFreeDet-FastSAM has been honored with both the best overall method and the best fast method awards in the model-free 2D detection track at BOP Challenge 2024 \cite{nguyen2025bopchallenge2024modelbased}.

\section{Conclusion}
\label{sec:conclusion}

In this paper, we have proposed GFreeDet, a model-free unseen object detection method leveraging Gaussian splatting and foundation models. Our unified Gaussian-based object reconstruction and template rendering pipeline enables effective foundation-model-driven segmentation/detection of unseen objects without CAD models. GFreeDet has established a strong baseline for the model-free unseen object detection benchmark on BOP-H3. 
We anticipate that this approach will inspire further research and applications in mixed reality.
However, despite advancements, the current approach, even in its fasted version, is not yet suitable for real-time MR systems. 
Future works will focus on enhancing the latency and robustness to facilitate real-time applications. Another promising direction would be to further reduce the onboarding cost required by the current setting, making it more accessible and user-friendly.

\section*{Acknowledgements}
We express our gratitude to the organizers of the BOP Challenge 2024 for their dedication and hard work in hosting the event.
This work is jointly supported by the National Natural Science Foundation of China under Grant No. 62406169, and the China Postdoctoral Science Foundation under Grant No. 2024M761673.

{\small
\bibliographystyle{alpha}
\bibliography{ref}
}


\end{document}